\pdfoutput=1

\documentclass[11pt]{article}

\usepackage[preprint]{acl}
\usepackage{multirow}
\usepackage{booktabs}
\usepackage{times}
\usepackage{latexsym}

\usepackage[T1]{fontenc}

\usepackage[utf8]{inputenc}
\usepackage{amsmath}

\usepackage{microtype}

\usepackage{inconsolata}

\usepackage{graphicx} 
\usepackage{natbib}  
\usepackage{caption} 

\usepackage{algorithm}
\usepackage{algorithmic}
\usepackage{amssymb}
\usepackage{pifont}
\definecolor{codeblue}{rgb}{0.25,0.5,0.5}
\definecolor{myblue}{rgb}{0.88,0.98,1}
\definecolor{mygreen}{rgb}{0.92, 1.0, 0.92}
\definecolor{myred}{rgb}{1, 0.9, 0.9}
\definecolor{mygray}{gray}{0.95}

\definecolor{Highlight}{HTML}{E8F8F5}
\definecolor{midgreen}{HTML}{589d62}
\definecolor{midblue}{HTML}{69a3f1}
\definecolor{darkgreen}{HTML}{146038}
\definecolor{darkblue}{HTML}{143b59}
\definecolor{hotpink}{RGB}{59, 115, 227}

\newcommand{\ours}{\textit{W2C }}
\title{World to Code: Multi-modal Data Generation via Self-Instructed Compositional Captioning and Filtering}

\author{
Jiacong Wang \textsuperscript{\rm1,2\thanks{\;\,These authors contributed equally to this work.}}
, Bohong Wu \textsuperscript{\rm2\footnotemark[1]}, Haiyong Jiang\textsuperscript{\rm1}, 
  Xun Zhou\textsuperscript{\rm2}, \\ \textbf{Xin Xiao}\textsuperscript{\rm2}, \textbf{Haoyuan Guo}\textsuperscript{\rm2}, \textbf{Jun Xiao}\textsuperscript{\rm1\thanks{\;\,Corresponding author.}} \\
\textsuperscript{\rm 1} School of Artificial Intelligence, University of Chinese Academy of Sciences \\
\textsuperscript{\rm 2} ByteDance Inc \\
  \texttt{wangjiacong20@mails.ucas.ac.cn}, 
    \texttt{\{haiyong.jiang,xiaojun\}@ucas.ac.cn} \\
        \texttt{\{bohongwu,guohaoyuan,xiaoxin.ddl\}@bytedance.com} \\
}

\begin{document}
\maketitle
\begin{abstract}
Recent advances in Vision-Language Models (VLMs) and the scarcity of high-quality multi-modal alignment data have inspired numerous researches on synthetic VLM data generation. 
The conventional norm in VLM data construction uses a mixture of specialists in caption and OCR, or stronger VLM APIs and expensive human annotation.
In this paper, we present World to Code (\textit{W2C}), a meticulously curated multi-modal data construction pipeline that organizes the final generation output into a Python code format. 
The pipeline leverages the VLM itself to extract cross-modal information via different prompts and filter the generated outputs again via a consistency filtering strategy. 
Experiments have demonstrated the high quality of \ours by improving various existing visual question answering and visual grounding benchmarks across different VLMs. Further analysis also demonstrates that the new code parsing ability of VLMs presents better cross-modal equivalence than the commonly used detail caption ability. Our code is available at \href{https://github.com/foundation-multimodal-models/World2Code}{https://github.com/foundation-multimodal-models/World2Code}.
\end{abstract}

\section{Introduction}
Fueled by the rapid development of Vision-Language Models (VLMs)~\cite{zhu2023minigpt,liu2024visual, team2023gemini,liu2024llavanext,dong2024internlm2} and Diffusion Models (DMs)~\cite{betker2023improving}, collecting detailed and concrete high-quality captions for each image becomes more and more urging. 
However, expensive and tedious human labeling for high-quality image-text pairs further incurs the necessity of a cheap and reliable data construction pipeline without human intervention.
\begin{figure*}
    \centering
    \includegraphics[width=1.0\linewidth]{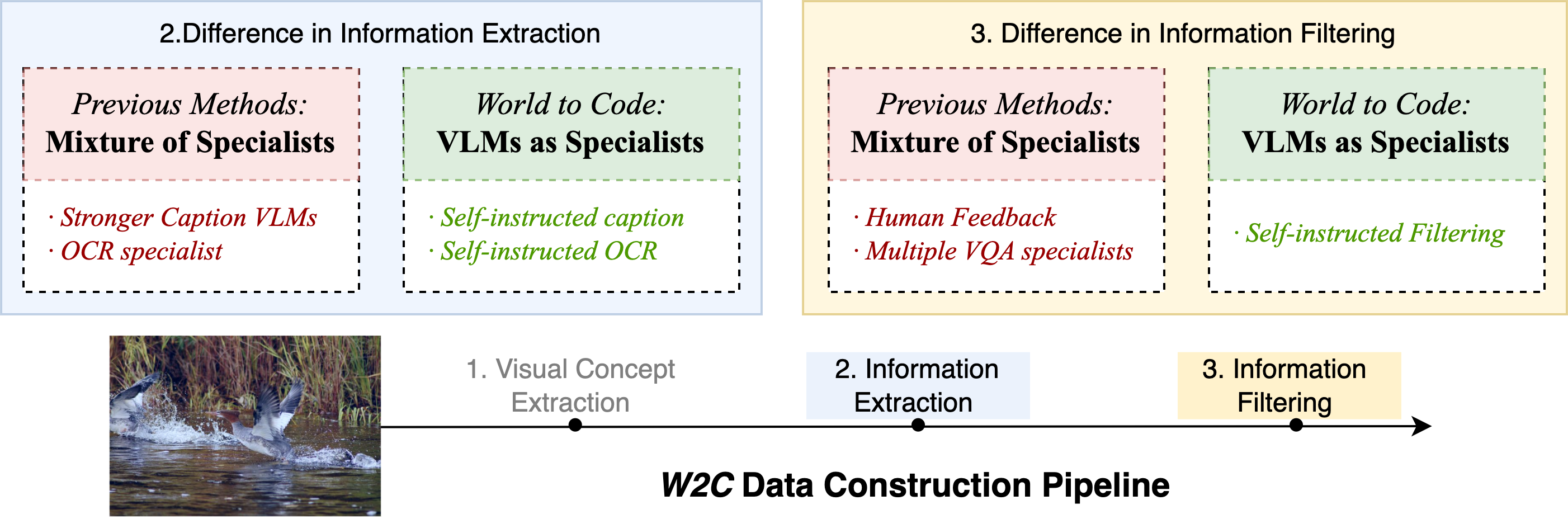}
    \caption{Overview of \ours and comparison of existing data construction pipelines. \ours differs from existing works by reducing the need for a mixture of specialists and expensive human annotations via self-instruct.}
    \label{fig:overview}
\end{figure*}

Related works on image-text data curation can be divided into two main streams. 
Distillation-based methods leverage closed-source commercial products (e.g., GPT-4V~\cite{achiam2023gpt}) with the state-of-the-art performance for image caption~\cite{chen2023sharegpt4v,li2024monkey,chen2024allava}.
Another line of work curates an image caption pipeline with existing VLMs to filter high-quality image-text for the training of better VLMs.
These methods usually combine open-source LLMs~\cite{touvron2023llama,touvron2023llama2,vicuna2023} and different visual specialists~\cite{li2023blip,huang2023tag2text,zong2023detrs,zhang2024recognize,fang2023eva,minderer2022simple,ren2024prompt,zhang2023gpt4roi} to endow existing VLMs with new abilities, e.g., pixel grounding in GLaMM~\cite{rasheed2023glamm}. 
However, the dependency on a mixture of specialists and human feedback in filtering noisy generations~\cite{wang2023all} makes it difficult to scale the generated data and automate the process. 

Recent progress shows that generated results of LLMs~\cite{wang2022self,li2023benchmarking} and VLMs~\cite{zhang2024unveiling} for prompts with similar meanings should be alike and we can help filter out noisy generated texts and captions by consistency checking among multiple prompt instructed results. 
In light of the above evidence, we present a self-instructed data construction pipeline, coined \ours. 
\ours autonomously extracts and articulates specific content from images, and enhances the reliability of the generated image captions by employing consistency filtering by assessing the outputs through multiple instructed prompt consistencies.
The overall pipeline reduces requested specialists and frees off expensive human feedback as shown in Figure~\ref{fig:overview}.
In addition, we leverage the idea from human-machine interaction and organize the model-generated responses into a Python code format, following Eureka~\cite{ma2023eureka} and Text2Reward~\cite{text2reward}. 
Experiments have shown that our proposed \ours can improve VLMs on various visual question-answering benchmarks. 
To be specific, \ours performs the best in 7 out of 9 VQA benchmarks on LLaVA-NeXT-7B, and 6 out of 9 VQA benchmarks on LLaVA-NeXT-13B. Furthermore, \ours also improves few-shot evaluations on two widely used VQA benchmarks including GQA and MME. 
Especially, on the 2-shot evaluation of GQA, the method achieves over 5 accuracy gains across different VLMs.

Our contribution is summarized in threefold:
\begin{itemize}
    \item We present the data pipeline of \ours, which proposes to generate and filter data all by existing VLMs themselves via self-instruct, significantly reducing the need for a mixture of specialists or expensive human annotations in conventional pipelines.
    
    \item The generated data of \ours presents comparable better performance on classical VQA benchmarks and consistently better performance on visual grounding benchmarks than ShareGPT4V.
    
    \item Further analysis presents that the new code parsing ability displays better cross-modality equivalence than the commonly used detail caption ability in presenting the details of an image.
\end{itemize}

\section{Related Work}
\paragraph{Vision Language Models} 
With the emergence of LLMs~\cite{chatgpt,achiam2023gpt,touvron2023llama,team2023gemini,jiang2024mixtral}, VLMs~\cite{zhu2023minigpt,zhang2023internlm,team2023gemini} have demonstrated exceptional capabilities in visual recognition and understanding, achieving remarkable results on various VLM benchmarks~\cite{singh2019towards,tito2021document,zhang2024unveiling,liu2023hidden,ying2024mmt,fu2024mme}. 
The seminal BLIP2~\cite{li2023blip} firstly introduces Q-Former to adapt encoded image features as potential language tokens for LLM-based caption prediction. 
Following works~\cite{liu2024llavanext,team2023gemini,dong2024internlm} improve the visual component by replacing VIT~\cite{dosovitskiy2020image} or scaling the input image resolution, while Zhu et al.~\cite{zhu2023minigpt} extends BLIP2 by employing emergent open-source LLMs~\cite{touvron2023llama,vicuna2023}, endowing current VLMs with significantly better instruction following and problem solving abilities. LLaVA/LLaVA-1.5~\cite{liu2024visual,liu2023improved} further remove Q-Former and point out that simple MLP projection layers present impressive performance in aligning image representation with LLMs.
Some works also highlight the importance of collecting high-quality cross-modal alignment data for improving the consistently scaling VLMs~\cite{bai2023qwen,wang2023all,li2023m}. 

\paragraph{Multi-modal Dataset Construction}
The scarcity of high-quality human-labeled data inspires the synthesis of cross-modal data~\cite{wang2024all,chen2023sharegpt4v,rasheed2023glamm,wang2023caption,li2024monkey,lu2023self,dong2024benchmarking,chen2024self}.
Among them, \citet{wang2023all} propose the AS-1B data generation pipeline and open-sourced high-quality dense captions on 1B images. GLaMM~\cite{rasheed2023glamm} further extends AS-1B by introducing about 10 specialists of different functionalities including grounding, tagging, and in-context learning. These specialists enable pixel-wise grounded dense captions for each image. However, the expensive human annotation required in AS-1B and the complicated construction pipeline in GLaMM have greatly limited the potential of data scaling. 
In this work, we try to answer whether synthetic data can improve VLMs on classical VQA benchmarks~\cite{fu2024mme,ying2024mmt,chen2024we} to avoid tedious data collection. 

Recent progress in synthetic data generation for LLMs~\cite{huang2023large,li2023benchmarking,wang2022self,wang2023self} shed light on the possibility of Multi-modal data construction by leveraging consistency in generation to filter invalid data. 
\citet{wang2022self} presents the consistent reasoning path generation demonstrating better performance in COT. 
\citet{li2023benchmarking} uses the generator-validator consistent data for training and can effectively improve LLMs on various tasks. 
\citet{zhang2024unveiling} further shows that the generator-validator consistency in most VLMs is prone to be correct.

\paragraph{Code Representation for Visual Tasks}

Code representations can formally encode various structure information in a scene.
Eureka~\cite{ma2023eureka} and Text2Reward~\cite{text2reward} parse a scene into Python codes and encourage LLMs to generate programmable dense rewards.
ViStruct~\cite{chen2023vistruct} takes the first step in visual code intelligence by decomposing the code-visual representation into multiple components including object recognition, object grounding, attribute detection, relation detection, and event detection. 
\citet{chen2023vistruct} further introduces a curriculum learning approach to endow VLMs with the aforementioned four abilities. 
However, the heavy dependency on supervised human-labeled datasets and the complicated curriculum learning pipeline limits its potential.
This work investigates an effective data-constructing pipeline based on code-vision representation.

\section{Method}
Our data construction pipeline shares some similarities with GLaMM~\cite{rasheed2023glamm}, where both methods focus on the region-level caption of the whole image. \ours further extend GLaMM to support generation-validation consistency filtering by exploring different organization formations of the labeled elements and present how VLMs boost themselves on basic multi-modal understanding tasks.

To make a comprehensive and systematic exposition of our \ours entire pipeline, the following will be divided into three parts for discussion:

(1) Visual Concepts Extraction in Section~\ref{section:visual concepts}, (2) Self-Instructed Information Extraction in Section~\ref{section:self taught}, (3) Information Filtering via Self Consistency in Section~\ref{section:filter}, (4) Structured formatting in Section~\ref{section:structure_formatting}. The overview of our construction pipeline is shown in Figure~\ref{fig:info_flow} and all the used instruct prompts are shown in Appendix~\ref{appendix:prompt}.

\begin{figure*}[t]
    \centering
    \includegraphics[width=1.0\linewidth]{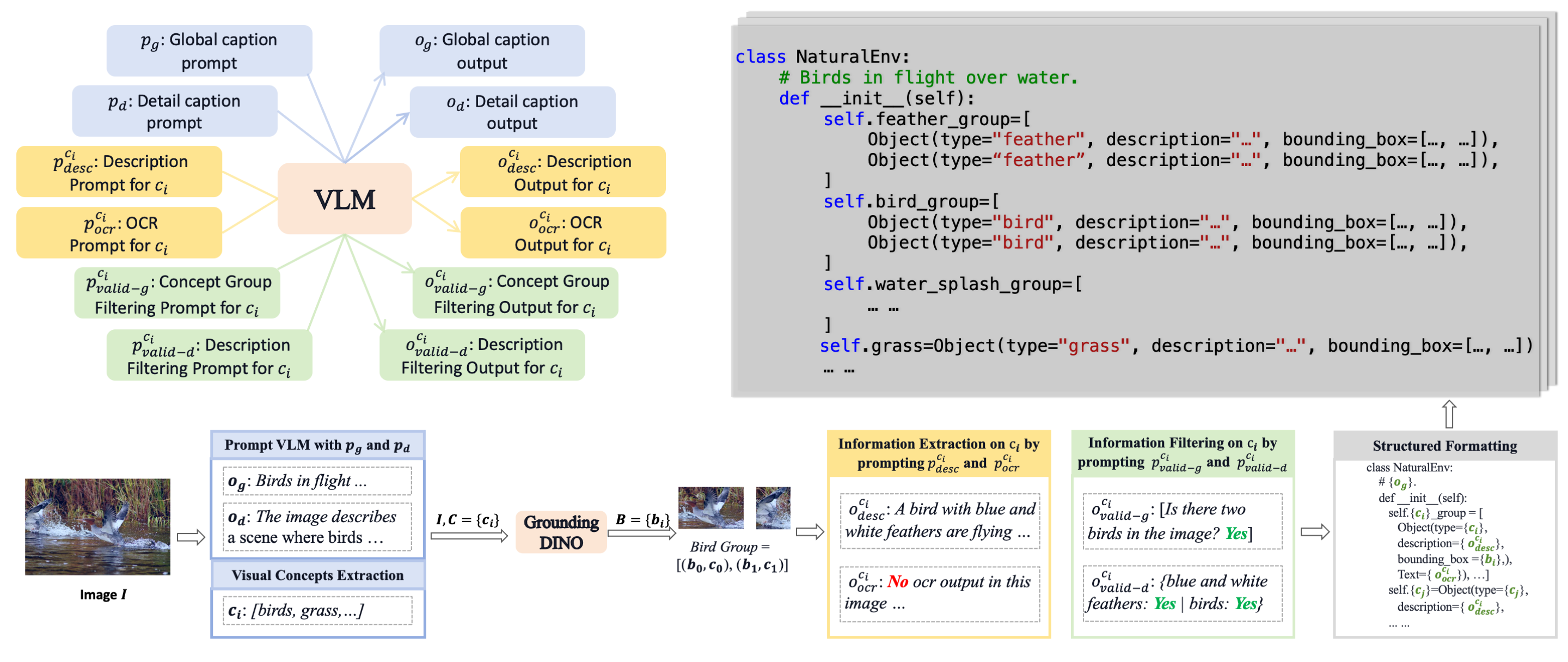}
    \caption{The data construction pipeline for \ours. Our pipeline utilizes both VLM and an object detector model to furnish structured data with region-specific awareness, detailed entity captions, and comprehensive global information. The VLM is iteratively invoked to generate the caption and perform consistency filtering to obtain high-quality data.
    The visual concepts set is obtained from the captions by the NLTK toolkit, $c_i$ here represents a visual concept from the set. The instruction prompts are all predefined templates.}
    \label{fig:info_flow}
\end{figure*}

\subsection{Visual Concepts Extraction} \label{section:visual concepts} 

To build a fully covered concept list for each image $I$ in images dataset $D_\text{raw}$, we prompt VLMs to generate both general captions (for a concise overview of the image) and detail captions (to bootstrap as many visual concepts as possible in the caption) using specific instruct prompts $p_{g}$ and $p_{d}$.
We use beam search to encourage the VLMs to provide as many visual concepts as possible to improve generation diversity.
The general captions and detail captions are obtained as follows:
\begin{equation}
o_{g}, o_{d} = f_{\text{VLM}}(I, p_{g}), f_{\text{VLM}}(I, p_{d}). 
\end{equation}
Since visual concepts are mainly composed of noun phrases, we employ the NLTK toolkit~\cite{bird2006nltk} to extract all noun phrases denoted as ${N}$ from $o_{g}$ and $o_{d}$. 

We use Grounding DINO to map extracted noun phrases to bounding boxes of the current image, where part of the false positive noun phrases are filtered as they fail to be mapped with corresponding areas in the image. Here we denote the filtered visual concepts as $\mathbf{C}$, and their corresponding bounding boxes as $\mathbf{B}$, which is formulated as follows:
\begin{equation}
    \mathbf{B}, \mathbf{C} = f_{\text{DINO}}(I, \mathbf{N}).
\end{equation}

\subsection{Self-Instructed Information Extraction} \label{section:self taught}
After obtaining visual concepts, we extract region-level captions and OCR information for cropped images of each concept bounding box, respectively.

\paragraph{Region-level Captions}  
We crop image $I$ for each visual concept $c_i \in \mathbf{C}$ with its corresponding bounding box $b_i \in \mathbf{B}$ to obtain a detailed caption and prompt the VLMs to provide a general caption centered on $c_i$. Additionally, to encourage the VLMs to offer more concrete details about the properties of $c_i$, we instruct the VLMs to include the color and material of $c_i$ in the caption. 
Denote the description prompt for the region-level caption as $p_{\text{desc}}(c_i)$ and the image cropped by $b_i$ as $I(b_i)$. The region-level caption for each visual concept $c_i$ is formulated as:
\begin{equation}
 o_{\text{desc}}(c_i) = f_{\text{VLM}}(p_{\text{desc}}(c_i), I(b_i))
\end{equation}

\paragraph{OCR information} Previous methods mainly use OCR tools~\cite{ocrtool,easyocr} to enhance the OCR capabilities. In contrast, \ours acquire the OCR information via an instructed prompt to guide VLMs for existing VLMs have the better capability in reading text in complex natural scenarios. Given the OCR instruct prompt $p_{ocr}(c_i)$, the OCR information in each cropped image by bounding box area $b_i$ with concept $c_i$ is formulated as follows:
\begin{equation}
 o_{\text{ocr}}(c_i) = f_{\text{VLM}}(p_{\text{ocr}}(c_i), I(b_i)) 
\end{equation}

\subsection{Information Filtering via Self Consistency} \label{section:filter}
Our consistency filtering strategy is inspired by the similar generator-validator consistency findings in ConBench~\cite{zhang2024unveiling}, where different instruct prompts may lead to in-consistent captions of visual concepts, and the highly consistent generations are prone to be correct ones. In this paper, we propose to filter the visual concepts via generation-validation consistency, where we change the region-level captions into multiple visual question answering problems for both counting filtering and caption reranking.

\paragraph{Counting Filtering via Consistency} Different from AS-1B, we introduce Grounding DINO in our construction process, which can naturally filter part of the plausible visual concepts as these concepts usually fail to find corresponding bounding boxes in the image. However, Grounding DINO introduces new challenges for counting problems, as visual concepts $c_i$ might be mapped to multiple boxes that have a large overlap due to inappropriately designed hyper-parameters. To prevent the effect by plausibly mapped $(b_i, c_i)$, we group all the $c_i$ that have the same name into $\tilde{\mathbf{C}}$ and calculate the existing times for each $\tilde{c}_i \in \tilde{\mathbf{C}}$ as $n_i$. We then merge all the boxes for each $\tilde{c}_i$ (which might contain multiple visual concepts with the same name) into $\tilde{\mathbf{B}}$, for a box $\tilde{b}_i \in \tilde{\mathbf{B}}$ we crop the image and prompt the VLMs to check whether the group element $\tilde{c}_i$ exist $n_i$ times in the image via instruct prompt $p_{\text{valid-g}}^{\tilde{c}_i}$:
\begin{equation}\label{eq:group_val}
o_{\text{valid-g}}(\tilde{c}_i) = f_{\text{VLM}}(p_{\text{valid-g}}(\tilde{c}_i,n_i), I(\tilde{b}_i)). 
\end{equation}

\paragraph{Caption Re-ranking via Consistency} To provide better region-level captions for a given visual concept, we use beam search to bootstrap multiple caption candidates. To select the best candidate, we again leverage the generator-validator consistency. 
Specifically, for each given visual concept $c_i$, we get a list of caption candidate $[o^{1}_{\text{desc}}(c_i), o^{2}_{\text{desc}}(c_i), ..., o^{b}_{\text{desc}}(c_i)]$. We use NLTK to parse these captions and collect all the visual concepts that are contained in these captions. Taking $n$ as the total number of extracted concepts in the captions of $c_i$, we get a new visual concept list denoted as $c^k_i \in \textbf{C}_\text{rank}$.
Following Equation~\ref{eq:group_val}, we prompt VLMs to check the existence of each extracted visual concept $c^k_i$ via instruct prompt $p_{\text{valid-c}}(c^k_i)$:
\begin{equation}
o_{\text{valid-c}}(c^k_i) = f_{\text{VLM}}(p_{\text{valid-c}}(c^k_i), I(\tilde{b}_i)) 
\end{equation}

We then manually design a scoring mechanism based on the validation result $o_{\text{valid-c}}(c^k_i)$. Specifically, for each caption that contains multiple extracted visual concepts, we assign each correct visual concept $o_{\text{valid-c}}(c^k_i)=\text{"Yes"}$ to score 1 and each hallucinated visual concept $o_{\text{valid-c}}(c^k_i)=\text{"No"}$ to -1. By accumulating the scores in each caption, we select the caption with the highest score in one beam as the final caption $o_{\text{desc}}(c_i)$ for the given visual concept $c_i$, which is supposed to be the most diverse and correct caption.

    \begin{algorithm}[tp]
      \caption{Data Construction and Consistency Filtering Pipeline}
      \label{code:baseline}
      \footnotesize
      \renewcommand{\algorithmicrequire}{\textbf{Input:}}
      \renewcommand{\algorithmicensure}{\textbf{Output:}}
      \begin{algorithmic}[1]
      \REQUIRE Image $I$ from dataset $D_\text{raw}$, Instruct Prompts: $p_\text{g}$, $p_\text{d}$, $p_{\text{desc}}$, $p_{\text{ocr}}$, $p_{\text{valid-g}}$, $p_{\text{valid-c}}$, VLM $f_{\text{VLM}}$, Grounding DINO $f_{\text{DINO}}$.
      \STATE Global Caption Generate. \\
      \quad $o_{g}, o_{d} = f_{\text{VLM}}(I, p_{g}), f_{\text{VLM}}(I, p_{d})$
      \STATE Visual Concepts Extraction. \\
      \quad$\mathbf{N} = NLTK(o_{g}, o_{d}),\quad \mathbf{B},\mathbf{C} = f_{\text{DINO}}(I, \mathbf{N}) $ \\
      \STATE Region-level Captions Generate.($c_i$ from $\mathbf{C}$,$b_i$ from $\mathbf{B}$) \\
      \quad$o_{\text{desc}}(c_i) = f_{\text{VLM}}(p_{\text{desc}}(c_i), I(b_i)) $ \\
      \STATE OCR information Extraction. \\
      \quad$o_{\text{ocr}}(c_i) = f_{\text{VLM}}(p_{\text{ocr}}(c_i), I(b_i))  $ \\
      \STATE Grouping Concepts and Boxes in $\mathbf{C}$ and $\mathbf{B}$. \\
      \quad $\tilde{c}_i \in \tilde{\mathbf{C}}$,\quad $\tilde{b}_i \in \tilde{\mathbf{B}}$
      \STATE Counting Filtering via Consistency. ($c^k_i \in \textbf{C}_\text{rank}$) \\
      \quad$o_{\text{valid-g}}(\tilde{c}_i) = f_{\text{VLM}}(p_{\text{valid-g}}(\tilde{c}_i,n_i), I(\tilde{b}_i))  $ \\
      \STATE Caption Re-ranking via Consistency. \\
      \quad$o_{\text{valid-c}}(c^k_i) = f_{\text{VLM}}(p_{\text{valid-c}}(c^k_i), I(\tilde{b}_i))   $ \\
     \STATE Rule-based Structured Formatting and Counting Filtering to get $D_{W2C}$. \\
      \quad$ $ \\
      \ENSURE \ours dataset $D_{W2C}$
      \end{algorithmic}
      \label{algo:v1}
    \end{algorithm}

\subsection{Structured Formatting and Filtering} \label{section:structure_formatting}
As shown in Figure~\ref{fig:info_flow}, we organize the structured information into code format to fully represent the region-level information of an image. Inspired by Eureka~\cite{ma2023eureka} and Text2Reward~\cite{text2reward}, we organize the information as a structured representation into the Python format due to its generality and conciseness. The organization is achieved by the following three rules.

\begin{itemize}
    \item One general caption $o_g$ of the whole image as the comments of each image Class.
    \item Each visual concept is an attribute for the image class. For each visual concept $c_i$, we get their corresponding bounding box $b_i$, caption $o_{\text{desc}}(c_i)$, and OCR information $o_{\text{ocr}}(c_i)$. Such visual concept is then organized as $\{\text{caption:} o_{\text{desc}}(c_i), \text{text:} o_{\text{ocr}}(c_i), \text{bbox:} b_i\}$.
    \item Grouping visual concepts with the same name. To make the representation code more concise, we group the visual concepts with the same name in a list $\tilde{c}_i\prime = [c_i^1, c_i^2, ...]$.
\end{itemize}

By integrating these rules, we get the final code representation of each image, which is then followed by the rule-based filtering strategy that filters out counting in-consistent samples.

In conclusion, by denoting the final dataset as $D_{W2C}$, the whole data construction pipeline is depicted in Algorithm~\ref{code:baseline}.

\section{Experiments}

\begin{table*}[ht]
\centering
\small
\begin{tabular}{l|ccccccccc}
\toprule
{Method} & {GQA} & {MME.} & {POPE} & {SQA$^I$}  & {MMS.} & {MMT.} & {Text.}  & {Doc.} & {Chart.}  \\

\midrule

\multicolumn{10}{c}{\small \em Low resolution setting}\\

\midrule

{LLaVA-1.5-7B}$^{*}$  &62.3 &1468 &\textbf{86.2} &68.2 &32.4 &48.6  
&\underline{47.6}&- &-\\
{\quad+ShareGPT4V}  &\textbf{63.4} &\textbf{1507}&\underline{86.0} & \underline{69.0}&\textbf{34.3} &\underline{49.3} 
& \textbf{47.9}&- &-\\
{\quad+\ours}  &\underline{62.8}&\underline{1503}&85.6 &\textbf{69.8} & \underline{33.5}&\textbf{49.4} 
&46.6&- &-\\
\midrule

{LLaVA-1.5-13B}$^{*}$ &63.7 & \textbf{1574}&85.7 &\underline{72.1} &33.5 & \underline{51.1}
 &\textbf{49.0}&- &-\\
{\quad+ShareGPT4V} &\textbf{64.0} &1537&\textbf{86.1} &72.0 & \underline{33.9}&50.9   
 & 48.8&- &-\\
{\quad+\ours} & \textbf{64.0}&\underline{1547} & 85.7&\textbf{72.6} &\textbf{36.1} &\textbf{51.7}  
&\underline{48.9}&- &-\\

\midrule
\multicolumn{10}{c}{\small \em High resolution setting}\\
\midrule

{LLaVA-NeXT-7B}&\textbf{64.2} &1473 &\underline{87.3} &67.9 &\underline{34.6} &48.2& \underline{63.9}  &\underline{75.4} &62.0  \\
{\quad+ShareGPT4V}$^{*}$ &64.0 &\underline{1513} &85.8 &\textbf{68.5} &33.7 &\underline{49.5} &\textbf{64.2} &75.1 &\underline{62.2}  \\
{\quad +\ours} &\textbf{64.2} & \textbf{1516}&\textbf{87.5} &\underline{68.3} &\textbf{35.8} &\textbf{50.1} &63.7&\textbf{76.5} & \textbf{63.0}   \\

\midrule

{LLaVA-NeXT-13B}&65.3&1545 &87.1 &70.1 &  \underline{37.2}&\underline{50.6}&\textbf{67.6} & 78.1&\textbf{66.2}  \\ 
{\quad+ShareGPT4V}$^{*}$ 
&65.3 &\underline{1574} & 87.1&70.1 &\textbf{37.5} &50.4  &\underline{67.0}&\underline{78.4} &63.8  \\
{\quad +\ours} &\textbf{65.5} &\textbf{1597} &\textbf{87.5} &\textbf{70.7} &37.1 &\textbf{51.4}&65.2 &\textbf{79.1} &\underline{65.6}  \\

\midrule

\end{tabular}
\caption{Visual Question Answering benchmarks of \ours on LLaVA1.5 and LLaVA-NeXT under different combination of IT datasets. The best results are \textbf{bold} and the second results are \underline{underlined}. $^{*}$: our reproduction of LLaVA-1.5 and LLaVA-Next, which achieves comparable performance with the original papers. $-$: LLaVA-1.5 does not support benchmarks that requires high input resolution. Abbreviations: SQA$^I$(ScienceQA), MMS.(MMStar), MMT.(MMT-Bench), Text.(TextVQA), Doc.(DocVQA), Chart.(ChartQA).}
\label{tab:1_main_result}
\end{table*}

\subsection{Experimental Setup}

\paragraph{Datasets} 
For the data construction pipeline, 
we strictly use the images in the ShareGPT4V dataset 
for our self-instructed approach validation in a fair comparison. 
Since the original ShareGPT4V dataset contains duplicate images, We remove the repeated images in the original 102K data and get about 87K original images. 
We follow the practice of LLaVA-1.5~\cite{liu2023improved} to adopt a two-stage training approach consisting of prompt tuning (PT) and instruct tuning (IT). 
For the experiments on low resolution setting, we follow the LLaVA-1.5 to use training dataset $\text{LLaVA}_\text{558k}$ for PT stage and $\text{LLaVA}_\text{665k}$ for IT stage on LLaVA-1.5 training stages.
As the specific mixture ratio details of the LLaVA-NeXT data were omitted, we directly utilized the entire training set from each of the following datasets in the IT stage, forming a mixture of datasets including:
$\text{LLaVA}_\text{665k}$~\cite{liu2023improved}, DocVQA~\cite{tito2021document}, ChartQA~\cite{masry2022chartqa} and ShareGPT4V~\cite{chen2023sharegpt4v} on high resolution setting. 

To comprehensively assess the effectiveness of our constructed dataset, we evaluate the model on widely adopted multi-modal benchmarks and grouding benchmarks, 
including TextVQA~\cite{singh2019towards} (without providing OCR tokens), DocVQA~\cite{tito2021document}, ChartQA~\cite{masry2022chartqa}, MME~\cite{fu2024mme}, MMT Bench~\cite{ying2024mmt}, MMStar~\cite{chen2024we}, ScienceQA~\cite{lu2022learn}, POPE~\cite{li2023evaluating}, 
GQA~\cite{hudson2019gqa}, RefCOCO~\cite{kazemzadeh2014referitgame}, RefCOCO+~\cite{mao2016generation} and RefCOCOg~\cite{mao2016generation}. 
These benchmarks provide a comprehensive assessment of multiple perspectives on multi-modal VLM performance. 
\paragraph{Implementation Details} 
In this paper, we employ two types of leading methods: LLaVA-1.5~\cite{liu2023improved} uses a CLIP-pretrained ViT-L/14~\cite{radford2021learning} as a vision encoder, a projector and an LLM, and LLaVA-NeXT~\cite{liu2024llavanext} increases the input image resolution by applying an adaptive image cropping strategy to concatenate all vision tokens.
To ensure a fair and comprehensive comparison Table~\ref{tab:1_main_result} and Table~\ref{tab:1_1_grounding_improve} present results both excluding and including the ShareGPT4V dataset, as well as results from the incorporation of our dataset. 
Table~\ref{tab:5_moresota}
We have reproduced LLaVA-NeXT with a learning rate of ViT to 1/10 of the base learning rate for the reason that LLaVA-NeXT only publishes their evaluation code. 
The learning rate for the PT stage is set to $1e^{-3}$ and the IT stage is set to $2e^{-5}$ for both Vicuna-7B and Vicuna-13B backbone LLM. We use 16 A100 for experiments on VLM training. We freeze the vision encoder during training on the LLaVA-1.5 and only freeze the vision encoder on the PT stage during training on the LLaVA-NEXT following the original paper. We show more training details in the Appendix~\ref{appendix:data}
\paragraph{Data Processing Details}
During the data construction pipeline, we employ NLTK~\cite{bird2006nltk} tool to extract noun phrases from the captions, and the resulting set of phrases is then post-processed using WordNet~\cite{miller1995wordnet} to remove duplicates and filter out inaccurately named entities. 
The total amount of final data after consistency filtering will not be completely consistent for different VLMs and we show the details in Appendix~\ref{appendix:data}.
The checkpoints of the VLM we used in our data processing are the original checkpoints of the official release. For LLaVA-1.5, which is not trained with the ShareGPT4V dataset, LLaVA-NEXT is trained with part of the ShareGPT4V dataset. The detailed GPU hours can be found in Appendix~\ref{appendix:detail} and we show the visualization of our \ours samples in Appendix~\ref{appendix:cases}.
\begin{table*}[ht]
\centering
\small
\begin{tabular}{l|cccccccc|c}
\toprule
{Method}  &\multicolumn{3}{c}{\textbf{RefCOCO}}   & \multicolumn{3}{c}{\textbf{RefCOCO+}}  & \multicolumn{2}{c|}{\textbf{RefCOCOg}} &\\
 & {test-a} & {test-b}& {val}   & {test-a} & {test-b} & {val} & {test} & {val} &Avg.\\
\midrule

\multicolumn{10}{c}{\small \em Low resolution setting}\\

\midrule

{LLaVA-1.5-7B}  & 86.8&\underline{72.9} & 80.0& 79.3& 60.7& 70.7&72.2 &72.2 &74.4\\
{\quad+ShareGPT4V}  & \underline{87.1}&72.7 &\underline{80.4} &\underline{79.5} &\underline{62.2} &\underline{71.5} & \underline{72.5}&72.2&\underline{74.8} \\
{\quad+\ours}  & \textbf{88.0}& \textbf{75.3}&\textbf{81.7} &\textbf{81.5} &\textbf{63.1} &\textbf{73.9} & \textbf{75.2}&\textbf{75.2} &\textbf{76.3}\\

\midrule

{LLaVA-1.5-13B} & 88.9& 75.3& 82.3& 82.4& 65.0&74.3 & 75.2&74.6&77.3 \\
{\quad+ShareGPT4V} &\underline{89.0}& \underline{75.6} &\underline{83.0} &\underline{82.7} & \underline{65.6}&\underline{75.7}&\underline{75.3}&\underline{75.0}&77.7 \\
{\quad+\ours} &\textbf{89.6} &\textbf{77.6}& \textbf{84.1}&\textbf{85.0} &\textbf{67.2} &\textbf{77.3} &\textbf{76.8}&\textbf{76.8}&\textbf{79.3 }\\

\midrule
\multicolumn{10}{c}{\small \em High resolution setting}\\
\midrule

{LLaVA-NeXT-7B} & \underline{89.9}& \underline{78.7}&\underline{84.8} &\underline{84.5}&\underline{68.7} &\underline{77.0} &\underline{79.4} & \underline{78.8}&80.2\\

{\quad +ShareGPT4V} &89.4 &76.8 &83.5 &82.1 &65.9 &75.5 &77.5 &77.6 &78.5\\
{\quad +\ours} &\textbf{90.9} &\textbf{81.3} &\textbf{86.4} &\textbf{85.8} &\textbf{70.5} &\textbf{79.5} &\textbf{80.7} &\textbf{80.5}&\textbf{82.0} \\

\midrule

{LLaVA-NeXT-13B} & \textbf{91.7}&\underline{81.9} & 86.3&86.2 & \underline{71.2}&79.5 & \underline{80.9}&\underline{80.8} &82.3\\

{\quad +ShareGPT4V} &\underline{91.5} &80.8 &\underline{86.5} &86.0 &71.1 &\underline{79.6} &79.6 &79.8&81.9 \\
{\quad +\ours} &91.1 &\textbf{83.6 }&\textbf{87.3} &\textbf{86.3} &\textbf{72.9} &\textbf{81.0} &\textbf{81.7 }&\textbf{81.3} &\textbf{83.2}\\

\midrule

\end{tabular}
\caption{Grounding benchmarks of \ours on LLaVA1.5 and LLaVA-NeXT under different combination of IT datasets. The best results are \textbf{bold} and the second results are \underline{underlined}.}
\label{tab:1_1_grounding_improve}
\end{table*}

\begin{table*}[ht]
\centering
\small
\setlength{\tabcolsep}{2.0pt}
\begin{tabular}{l|cccccccc}
\toprule
{Method} & {DocVQA} & {ChartQA} & {POPE}   & {MMStar} & {MMT-Bench}   &RefCOCO${_{val}}$  &   RefCOOC+${_{val}}$ &  RefCOCOg${_{val}}$ \\

\midrule

{LLaVA-NeXT-7B}$^{*}$  &75.4 &62.0 &87.3 &34.6 &48.2  
&84.8&77.0 &78.8\\
{\quad+ShareGPT4V}  &75.1 &62.2  & 85.8&33.8 &49.5 
&83.5&77.3 &77.6 \\
{\quad+ALLaVA} & \textbf{76.5} & \textbf{63.0} &87.3 &33.2 &\textbf{50.4} &\underline{85.5} 
&77.0 &\underline{79.9} \\
{\quad+Monkey}& 76.4 &62.5 &\textbf{87.5}  &\underline{35.7} &49.6 &85.0 &\underline{77.4} 
&78.2 \\
{\quad+\ours}  &\textbf{76.5}&\textbf{63.0}&\textbf{87.5} &\textbf{35.8} & \underline{50.1}&\textbf{86.4} 
&\textbf{79.5}&\textbf{80.5}  \\
\midrule

\end{tabular}
\caption{Visual Question Answering benchmarks and Grouding benchmarks on LLaVA-NeXT-7B under more combination of SOTA IT dataset methods. The best results are \textbf{bold} and the second results are \underline{underlined}. $^{*}$: our reproduction of 
LLaVA-Next, which achieves comparable performance with the original papers. To ensure a fair comparison, we randomly selected an equal amount of corresponding data from each dataset for this analysis.}
\label{tab:5_moresota}
\end{table*}

\begin{table*}[ht]
\centering
\small
\setlength{\tabcolsep}{4.0pt}\begin{tabular}{lc|ccccccc}
\toprule
{Method} & format & {MMT-Bench}  & {\ DocVQA \ } & {\ TextVQA \ }& {$\text{RefCOCO}_\text{val}$} & {$\text{RefCOCO+}_\text{val}$} &  {$\text{RefCOCOg}_\text{val}$}     \\

\midrule

{LLaVA-NeXT-7B} & \textit{single} &49.2 &75.4 & \textbf{63.8}& 85.4& 78.5&79.5  \\

{LLaVA-NeXT-7B} & \textit{multi} &48.8 &72.0 & 61.4&82.4& 73.8& 76.8 \\
{LLaVA-NeXT-7B} & \textit{code} &\textbf{50.1} &\textbf{76.5} & 63.7&\textbf{86.4} & \textbf{79.5}&\textbf{80.5}  \\

\midrule



\end{tabular}
\caption{Ablation study of \ours on using different data organization format. \textit{single/multi/code}: constructed data are organized in single-round conversations/multi-round conversations/python code format.}
\label{tab:2_code_format}
\end{table*}

\subsection{Main Results}
\paragraph{Effectiveness of \ours data improve various VLMs in Visual Question Answering benchmarks}
We show a quantitative comparison results of the trained VLMs with and without the ShareGPT4V dataset, as well as \ours for replacement of the ShareGPT4V during the IT training stage in Table~\ref{tab:1_main_result}.
\ours consistently improves the performance on different settings in both LLaVA-1.5 and LLaVA-NeXT. Especially, in the high resolution setting, our \ours presents impressive performance improvement on multi-modal visual understanding benchmarks such as MMT Bench, MMStar, and MME. 
Specifically, \ours can bring improvement in 7 out of 9 benchmarks on LLaVA-NeXT-7B and 6 out of 9 on LLaVA-NeXT-13B. Especially, on LLaVA-NeXT-13B, \ours improves DocVQA by 0.7 ANLS, ChartQA by 1.8 accuracy, MMT Bench by 0.8 accuracy and MME by 23 points compared to the reproduction results of LLaVA-NeXT. More benchmarks results are shown in~\ref{appendix:morebench}.
\paragraph{\ours data show impressive performance on Grounding benchmarks}
We present the performance of the VLMs 
on Grounding benchmarks in Table~\ref{tab:1_1_grounding_improve}. The task of referential expression comprehension necessitates that the model accurately identifies and localizes the object described.
Our models demonstrate their exceptional capability for detailed image recognition and localization by undergoing evaluation across various referential expression comprehension benchmarks, including RefCOCO, RefCOCO+, and RefCOCOg. 
Benefit from the entity-enteric generation of local captions and the presence of local bounding box information, our model achieved an average improvement of 1.5/1.6 average IoU on LLaVA-1.5 7B/13B and 3.5/1.3 average IoU on LLaVA-NeXT-7B/13B.
{\paragraph{Comparison results of more data generation methods and \ours on LLaVA-NeXT-7B model under different benchmarks.} We show more quantitative results on the LLaVA-NeXT-7B baseline, employing more data generation methods (ALLaVA and Monkey) that utilize the GPT API for data annotation. To ensure a fair comparison, we randomly selected an equal amount of corresponding data from each dataset. We reported on representative Visual Question Answering and Grounding benchmarks and achieved the best outcomes in 7 out of 8 benchmarks. \ours still gets comparable results compared to ALLaVA and gets better results on Grounding benchmarks.}

\begin{table*}[ht]
\centering
\small
\setlength{\tabcolsep}{2.0pt}
\begin{tabular}{lcc|ccccccc}
\toprule

{Method} & \textit{re-ranking} & \textit{counting}& {MMT-Bench} & {\ DocVQA \ } & {\ TextVQA \ }& {$\text{RefCOCO}_\text{val}$} & {$\text{RefCOCO+}_\text{val}$} & {$\text{RefCOCOg}_\text{val}$}    \\

\midrule

{LLaVA-NeXT-7B} & & &\textbf{50.3} & 75.5& 62.7& \textbf{86.6}&79.0  & 79.7 \\
{LLaVA-NeXT-7B} &&\checkmark &49.4 & 76.3&63.4 &86.1 & 78.5 & 80.4  \\
{LLaVA-NeXT-7B} & \checkmark&&49.4 & 75.3&63.2& 86.5& 79.2 & 79.7 \\
{LLaVA-NeXT-7B} &\checkmark &\checkmark &50.1 &\textbf{76.5} & \textbf{63.7}&86.4 & \textbf{79.5}&\textbf{80.5} \\

\midrule

\end{tabular}
\caption{Ablation study of \ours when combined the different consistency filtering strategy. \textit{re-ranking}: caption re-ranking. \textit{counting}: counting filtering.}
\label{tab:3_filter}
\end{table*}

\begin{table}[tp]
\setlength{\tabcolsep}{5.5pt}
\centering
\small
\begin{tabular}{l|cccc}
\toprule
\multirow{2}{*}{Method} & \multicolumn{2}{c}{GQA} & \multicolumn{2}{c}{MME} \\

 & {2-shot} & {4-shot} & {2-shot} & {4-shot} \\

\midrule
\multicolumn{1}{l}{\em LLaVA-1.5-7B}\\

{detail caption} & 34.79 & 39.67 & 1136 & 1098   \\

{code parsing} & \textbf{41.06} & \textbf{43.40} & \textbf{1139} & \textbf{1169}  \\

\midrule
\em LLaVA-1.5-13B & & & \\
detail caption & 34.00 & 40.87 & 1192 & 1170 \\
code parsing & \textbf{39.12} & \textbf{43.70} & \textbf{1199} & \textbf{1224} \\

\midrule
{\em LLaVA-NeXT-7B} & & &\\
detail caption & 34.89 & 40.70 & \textbf{1174} & 1105  \\
code parsing & \textbf{40.07} & \textbf{45.07} & 1154 & \textbf{1189} \\

\midrule
{\em LLaVA-NeXT-13B} & & &\\
detail caption & 31.63 & 40.07 & \textbf{1193} & 1127  \\
code parsing & \textbf{39.80} & \textbf{42.83} & 1151 & \textbf{1190} \\
\midrule

\end{tabular}
\caption{Comparison between detail caption and code parsing ability in few-shot evaluations on MME and GQA without referring to the image.}
\label{tab:few_shot_code_eval}
\end{table}

\subsection{Ablation Studies}
Our results show advantageous performance in Table~\ref{tab:1_main_result} and Table~\ref{tab:1_1_grounding_improve}, but our analysis of these results shows the limitations of the base model's OCR capability on LLaVA-1.5. 
We proceed with further ablation studies on LLaVA-Next-7B for the constraints on resources, which optimally demonstrate the full benefits of our pipeline and consistency filtering in a comprehensive manner. 
\paragraph{Organizing data into the python code format presents better performance} We discussed in Section~\ref{section:self taught} the strengths of choosing the code format for the representation of structured data. In
Table~\ref{tab:2_code_format}, we quantitatively compare our data format with a single-round dialogue format and a multi-round dialogue format.
By using the python code as data construction format, we observe improved performance in both visual grounding benchmarks and visual question answer benchmarks on LLaVA-NeXT-7B. Especially, we improved the MMT-Bench by 0.9/1.3 accuracy and DocVQA by 1.1/4.5 ANLS compared to the \textit{single/multi} data format.
\paragraph{Filtering introduces better downstream benchmarks performance}
We show the ablation of different consistency filtering choices in Table~\ref{tab:3_filter}. 
Similarly, the performance of LLaVA-NeXT-7B on the both visual grounding benchmarks and visual question answering benchmarks highlights the effectiveness and necessity of our consistency filtering approaches. When two filtering strategies are combined, we achieve the best performance by improving DocVQA with 1.0 ANLS, TextVQA with 1.0 accuracy, RefCOCO+$_{val}$ with 0.5 IOU and RefCOCOg$_{val}$ with 0.8 IOU. We also achieve comparable results on MMT-Bench and RefCOCO$_{val}$ with little performance degradation.

\subsection{Code Parsing Ability Evaluation}

We further present better cross-modality equivalence between image and text brought by the new code parsing ability. An ideal caption of the image should enable the ability to question without referring to the image. Therefore, we compare the quality of the code output and widely used detail caption output in the ability to handle downstream tasks via in-context learning on the same Large Language Model.

\paragraph{Experimental Setting} We conduct experiments on both LLaVA-1.5-7B/13B and LLaVA-NeXT-7B/13B on two widely used Visual Question Answering benchmarks, including GQA and the perception subset of MME. Due to the support of 32k long context and satisfying performance in the open-source community, we use Qwen-1.5-14B~\cite{bai2023qwen,qwen1.5} as the problem-solving LLM, and prompt it with few shot inputs. Each shot can be represented as a combination of $\{\text{description, question, answer}\}$. For the detail caption output, we use the models trained with both the original dataset and the ShareGPT4V dataset to improve their detail caption abilities. For the code parsing output, we replace ShareGPT4V with our proposed \ours dataset.
\paragraph{The code parsing ability of VLMs presents much better few-shot performance.} From Table~\ref{tab:few_shot_code_eval}, the code parsing output shows significant improvement when compared with using the detail caption output. On the binary classification task for the visual perception subset of MME, the code parsing ability achieves comparable or better performance in various settings. On the free generation VQA task, GQA, using the code parsing output can bring clear accuracy gain across different model size and architectures. Especially, on the 2-shot evaluation of GQA on LLaVA-NEXT-13B, the code parsing output by model trained with \ours achieves 8.2 accuracy improvement compared to baseline, indicating that the code-parsing ability present improved performance in presenting the details of one image. More benchmarks results are shown in~\ref{appendix:moreshot}.

\section{Conclusion}
This paper presents \ours, an enhanced data construction pipeline that only leverages existing VLMs themselves for detail and compositional captions for an image, which is further organized in Python code format. We present that existing VLMs can improve themselves on the understanding benchmarks in various scenarios, significantly reducing the need for a mix of visual specialists and heavy human annotations. Moreover, additional experiments show that the new code parsing ability of VLMs presents better capability in fully describing the image, with notable improvement in the few-shot evaluation on downstream tasks when the raw images are not provided. Our proposed \ours not only enhances the original capabilities on the widely used multi-modal understanding benchmarks but also endows existing VLMs with detailed and executable multi-modal parsing ability.


\section{Limitation}
Despite the advancements in improved multi-modal understanding benchmarks and new code parsing ability, \ours can be further improved in some aspects. 
\begin{itemize}
    \item In this paper, we directly use the ShareGPT4V dataset images for a fair comparison with ShareGPT4V. However, it contains fewer OCR-centric images, limiting the final performance. Further investigation could be taken in studying the performance of \ours on more distribution of unlabeled datasets. 
    \item The experiments are mainly conducted on the SOTA open-source VLM structures, i.e., the LLaVA series which use MLP projectors for multi-modal alignment. The effectiveness of \ours can be further investigated on other VLM structures. 
\end{itemize}

Given the promising performance of \ours on evaluation benchmarks, we would like to explore a more high-quality and diverse data generation pipeline in future investigations.

\paragraph{Acknowledgments.} This work is partially supported by the National Natural Science Foundation of China (U21A20515, 62476262, 62102393, 62206263, 62271467, 2306297, 62306296), Beijing Natural Science Foundation (4242053, L242096), China Postdoctoral Science Foundation (2022T150639) and the Fundamental Research Funds for the Central Universities.

\bibliography{main}

\clearpage
\appendix
\section{Prompt Templates for \ours data construction pipeline}
\subsection{Prompt Templates} \ours data construction pipeline calls the VLMs repeatedly by using different prompts. We guide the VLMs to accurately answer questions by designing universal prompt templates, thus ensuring better compliance with instruction. All the prompts are shown in Table~\ref{tab:prompt}.
\begin{table*}[h]
\centering
\small
\begin{tabular}{l p{10.5cm} }
\toprule
{Stage} & Prompt \\

\midrule
\multicolumn{2}{l}{\bf \textit{Prompt for Caption}}  \\
\midrule
$\text{Global Caption}-p_g$& Please provide a simple sentence that describes this image accurately. \\
$\text{Detail Caption}-p_d$& Please describe all the visual concepts in the image in detail, but use concise words with no more than 120 words. \\
\midrule
\multicolumn{2}{l}{\bf \textit{Prompt for Self-Instructed Concept-targeted Captions}}  \\
\midrule
$\text{Compositional Caption}-p_{\text{desc}}$& From the image, provide one sentence that describes \{e\} (you should try your best to include attributes like shape, color or material), especially, using \{e\} as the beginning of your answer. \\


$\text{OCR Extract}-p_{\text{ocr}}$ & List all the text in the image, answer with the ocr tokens only, and answer 'No' with one word if there isn't any.  \\

\midrule
\multicolumn{2}{l}{\bf \textit{Prompt for Consistency Filtering}}  \\
\midrule
$\text{Caption Re-ranking}-p_{\text{valid-c}}$& Is '\{e\}' a valid and visible visual concept in the image? Answer yes or no with only one single word. \\
$\text{Counting Group Filtering}-p_{\text{valid-g}}$ & Is there \{parse times\} or more \{group key\} in the image? Answer yes or no with a single word. \\
\midrule 
\multicolumn{2}{l}{\bf \textit{Symbol Explanation}}  \\
\midrule 

\{e\} & means an entity in the final detected entity list of this image.\\
\{parse times\} & means the number of times an entity appears in the entity list of this image. \\
\{group key\} & means the entity name corresponding to parse times in the entity list of this image. \\
\midrule 

\end{tabular}
\caption{Prompt for \ours data construction pipeline.}
\label{tab:prompt}
\end{table*}

\label{appendix:prompt}

\section{More experiments of \ours.}
\subsection{More Visual Question Answering Benchmarks} We show more Visual Question Answering benchmarks of \ours on LLaVA-NeXT-7B/13B under different combination of IT datasets in Table~\ref{tab:6_morebench}. The \ours method consistently demonstrates superior experimental results.
\begin{table}[tp]
\setlength{\tabcolsep}{5.5pt}
\centering
\small
\begin{tabular}{l|cc}
\toprule
{Method} & {InfoVQA} & {MMMU} \\

\midrule
{LLaVA-NeXT-7B} &26.3&35.8\\
{\quad+ShareGPT4V} & 27.4 & 35.0   \\
{\quad+\ours} & \textbf{27.8} & \textbf{36.2}  \\
\midrule
{LLaVA-NeXT-13B} & 29.7&35.3\\
{\quad+ShareGPT4V} & 29.9 & 34.8   \\
{\quad+\ours} & \textbf{30.4} & \textbf{35.3}  \\
\midrule

\end{tabular}
\caption{More Visual Question Answering benchmarks of \ours on LLaVA-NeXT-7B/13B under different combination of IT datasets. The best results are \textbf{bold}.}
\label{tab:6_morebench}
\end{table}

\label{appendix:morebench}
\subsection{Code Parsing Ability Evaluation} We have added an analysis of in-context learning for two representative datasets in Table~\ref{tab:more_2shot}: MMStar and RefCOCOg.
It's important to note that although we report the in-context learning results on RefCOCOg val set under the same settings, comparing these two types of outputs for grounding tasks is not practically meaningful. This is because when we instruct the \ours-trained model to output in detailed caption format, the captions do not usually contain specific box information like [x1,y1,x2,y2]. This leads to a low IoU score for in-context learning with 2/4 shot detailed captions. However, when outputting in code format, the model does predict box information, which accounts for the significant difference in results on RefCOCOg.
\begin{table}[tp]
\setlength{\tabcolsep}{5.5pt}
\centering
\small
\begin{tabular}{l|cccc}
\toprule
{Method} & \multicolumn{2}{c}{MMStar} & \multicolumn{2}{c}{RefCOCOg} \\

 & {2-shot} & {4-shot} & {2-shot} & {4-shot} \\

\midrule
{\em LLaVA-NeXT-7B} & & &\\
detail caption & 33.60 & 33.47 & 11.49& 19.32  \\
code parsing & \textbf{35.13} & \textbf{36.00} & \textbf{49.83} & \textbf{51.02} \\

\midrule

\end{tabular}
\caption{Comparison between detail caption and code parsing ability in few-shot evaluations on MMStar and RefCOCOg without referring to the image on LLaVA-NeXT-7B.}
\label{tab:more_2shot}
\end{table}

\label{appendix:moreshot}

\section{Implementation Details for \ours experiments}
\subsection{Dataset Details}
All the creators or original owners of assets used in the paper are credited properly, and the license and terms of use are explicitly mentioned and are respected properly. All datasets we use are from internet open-source datasets under CC-BY licenses and are cited properly. 
\label{appendix:data}
\paragraph{Data Construction Pipeline Details} We incorporate images from the open-source ShareGPT4V dataset, totaling approximately 87K images. 
For the VLMs in our data construction pipeline, we directly use the official release checkpoints including LLaVA-1.5 and LLaVA-NeXT.

For the cost of our data construction pipeline, we use about 1/1.5 day on 32 A100s GPU for LLaVA-1.5 and about 2/3 days on 48 A100s GPU for LLaVA-NeXT.
For the data obtained by \ours pipeline, we get 34K from LLaVA-1.5-7B, 33K from LLaVA-1.5-13B, 37K from LLaVA-NeXT-7B, and 29K from LLaVA-NeXT-13B. 
The reasons for the inconsistency in the amount of data are multifaceted. On the one hand, a minor portion of the data was discarded due to improper handling of anomalous data throughout the processing stage. On the other hand, a significant amount of data was eliminated during the consistency filtering stage owing to inconsistencies detected by the VLMs. Additionally, the generative capabilities of various VLMs vary, and the inherent randomness within VLMs themselves also contributes to these inconsistencies.

\paragraph{Training Details} During the training of VLMs, we use different dataset combinations. 
We utilize the original paper's open-source dataset during both the PT and IT training stages for LLaVA-1.5. 
In contrast, for the training of LLaVA-NeXT, the lack of disclosure regarding the specific details of the IT stage, we trained using all training set from $\text{LLaVA}_\text{665k}$~\cite{liu2023improved}, DocVQA~\cite{tito2021document}, ChartQA~\cite{masry2022chartqa} and ShareGPT4V~\cite{chen2023sharegpt4v}. Furthermore, by aligning our dataset with that of the original study, we achieved comparable experimental results.
We use the CLIP-pretrained ViT-L/14~\cite{radford2021learning} as a vision encoder, which input resolution is 336$\times$336.
We freeze the vision encoder during training on the LLaVA-1.5 and only freeze the vision encoder on the PT stage during training on the LLaVA-NEXT following the original paper. 
The experiments of VLM training are all conducted on 16 A100 GPUs.
\subsection{Implementation Details of our Pipeline}
\label{appendix:detail}
We employ beam search to fully leverage the powerful language generation capabilities and extensive knowledge base of VLM. This approach enables the generation of an increased number of captions, assisting us in acquiring a broader set of visual concept candidates. Due to the limitation of GPU memory, we set the generation beam to 8 on LLaVA-1.5 and 4 on LLaVA-Next.
The learning rate for the PT stage is set to $1e^{-3}$ and the IT stage is set to $2e^{-5}$ for both Vicuna-7B and Vicuna-13B backbone LLM. We set the warmup ratio to 0.03, the PT stage batch size is set to 256 and the IT stage batch size is set to 128. We use model max length 2048 on LLaVA-1.5 and 4096 on LLaVA-Next for its high resolution setting.

\subsection{Data Example}
\begin{figure*}
    \centering
    \includegraphics[width=0.95\linewidth]{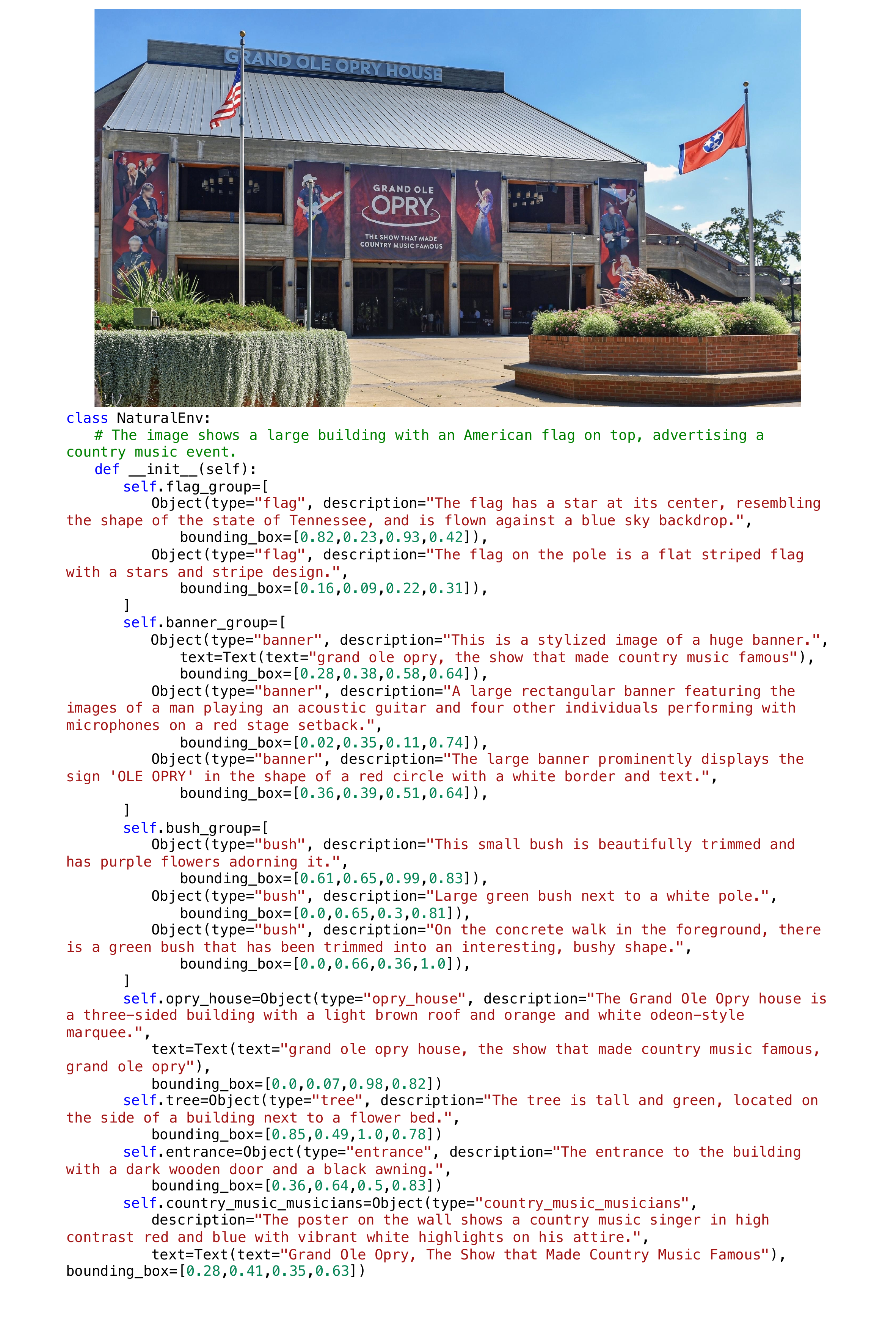}
    \caption{Visualization of one \ours sample with OCR information.}
    \label{fig:case1}
\end{figure*}

\begin{figure*}
    \centering
    \includegraphics[width=0.95\linewidth]{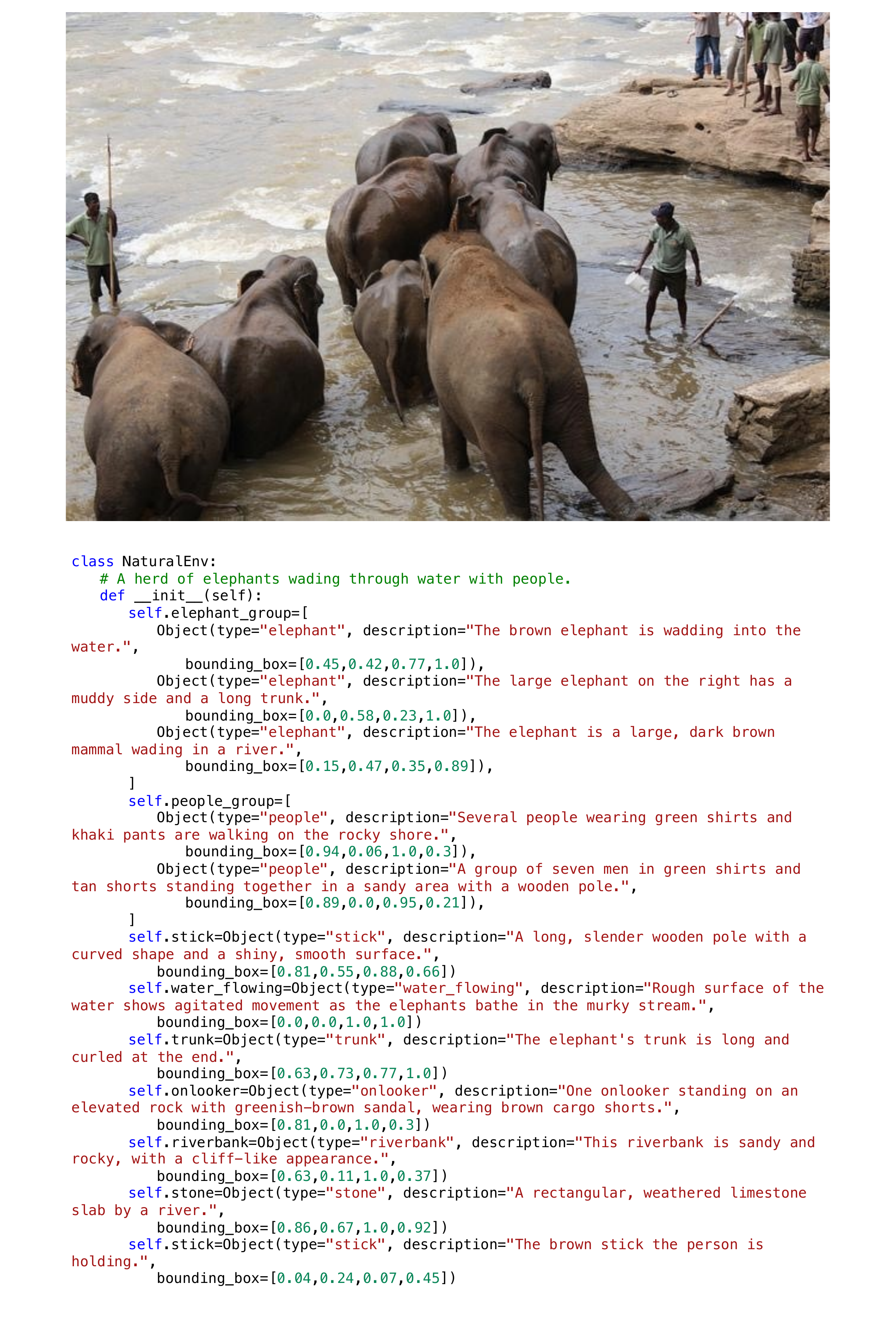}
    \caption{Visualization of one \ours sample without OCR information.}
    \label{fig:case2}
\end{figure*}

In Figure~\ref{fig:case1} and Figure~\ref{fig:case2}, we present images from the ShareGPT4V dataset alongside the corresponding annotations we constructed by \ours. As shown in these images, the annotations generated entirely by the VLMs accurately describe both the global captions and the detailed captions of local entities within specific areas. Additionally, the OCR text is also encapsulated within the corresponding frames. For multiple entities present in the images, a display of group merging is also conducted.

\label{appendix:cases}

\end{document}